\documentclass[10pt,twocolumn,letterpaper]{article}
\usepackage{cvpr}
\usepackage[ruled,vlined]{algorithm2e}
\definecolor{cvprblue}{rgb}{0.21,0.49,0.74}
\usepackage[accsupp]{axessibility}
\usepackage{xcolor}

\definecolor{cvprpink}{RGB}{220, 70, 130}
\usepackage{multirow}
\usepackage[table]{xcolor}
\usepackage{array}  
\newcommand{\g}{\cellcolor{gray!20}}
\usepackage[pagebackref,breaklinks,colorlinks,allcolors=cvprblue]{hyperref}
\usepackage{hyperref}
\def\paperID{} % *** Enter the Paper ID here
\def\confName{CVPR}
\def\confYear{2026}

\title{TTP: Test-Time Padding for Adversarial Detection and Robust Adaptation on Vision-Language Models}

\author{
  Zhiwei Li\textsuperscript{1,2}\thanks{Equal contribution.}
  \quad Yitian Pang\textsuperscript{3}\footnotemark[1]
  \quad Weining Wang\textsuperscript{2}
  \quad Zhenan Sun\textsuperscript{1,2}
  \quad Qi Li\textsuperscript{1,2}\thanks{Corresponding author.} \\
  \textsuperscript{1}NLPR \& MAIS, Institute of Automation, Chinese Academy of Sciences \\
  \textsuperscript{2}University of Chinese Academy of Sciences\quad 
  \textsuperscript{3}Department of Automation, Tsinghua University \\
  {\small
  lizhiwei2023@ia.ac.cn, pangyt23@mails.tsinghua.edu.cn, \{weining.wang, znsun, qli\}@nlpr.ia.ac.cn}
}

\begin{document}
\maketitle % 生成标题
% {\footnotesize *Equal contribution.}
% \renewcommand{\thefootnote}{*}

% \footnotetext{Equal contribution.}
% \renewcommand{\thefootnote}{\arabic{footnote}}  % 恢复原脚注样式
% 摘要（严格遵循用户提供的TTP相关内容，突出问题、方法、核心结果）
\begin{abstract}
Vision-Language Models (VLMs), such as CLIP, have achieved impressive zero-shot recognition performance but remain highly susceptible to adversarial perturbations, posing significant risks in safety-critical scenarios. Previous training-time defenses rely on adversarial fine-tuning, which requires labeled data and costly retraining, while existing test-time strategies fail to reliably distinguish between clean and adversarial inputs, thereby preventing both adversarial robustness and clean accuracy from reaching their optimum. To address these limitations, we propose Test-Time Padding (TTP), a lightweight defense framework that performs adversarial detection followed by targeted adaptation at inference. TTP identifies adversarial inputs via the cosine similarity shift between CLIP feature embeddings computed before and after spatial padding, yielding a universal threshold for reliable detection across architectures and datasets. For detected adversarial cases, TTP employs trainable padding to restore disrupted attention patterns, coupled with a similarity-aware ensemble strategy for a more robust final prediction. For clean inputs, TTP leaves them unchanged by default or optionally integrates existing test-time adaptation techniques for further accuracy gains. Comprehensive experiments on diverse CLIP backbones and fine-grained benchmarks show that TTP consistently surpasses state-of-the-art test-time defenses, delivering substantial improvements in adversarial robustness without compromising clean accuracy. Our code is available at \url{https://github.com/lizhiwei23/TTP}.
% The code for this paper will be released soon.
\end{abstract}
% 1. 引言（严格遵循用户提供的内容，按CVPR逻辑：领域背景→现有问题→TTP方法提出→贡献点）
\section{Introduction}
\label{sec:intro}
\begin{figure}[t]
  \centering
   \includegraphics[width=0.98\linewidth]{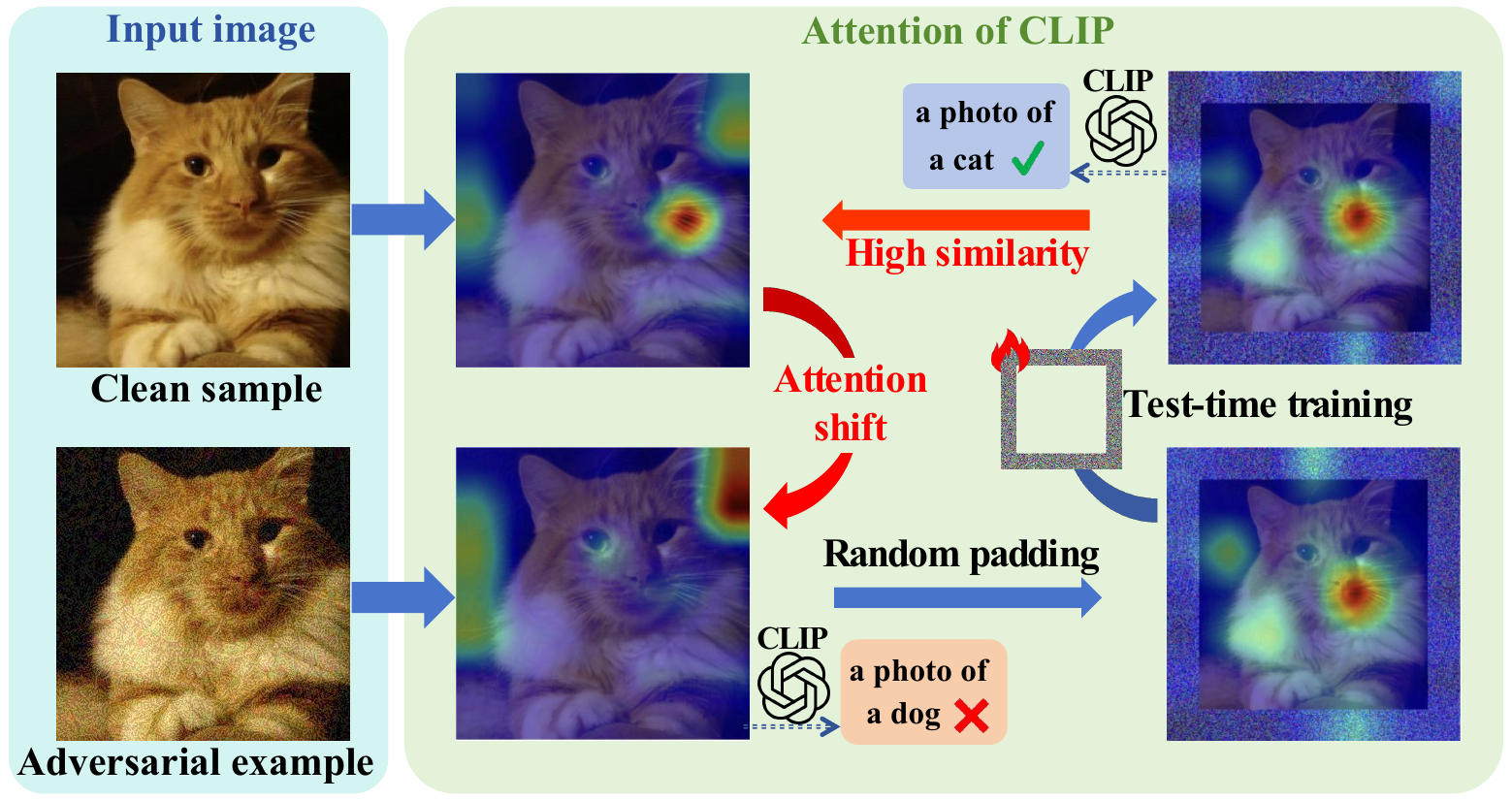}
    \caption{Visualization of attention maps for clean sample, adversarially perturbed sample, randomly padded sample, and samples processed with trainable test-time padding. The adversarial attack causes a noticeable shift in attention, leading to incorrect predictions. Applying random padding helps restore the original attention focus, while trainable padding further refines the attention to the correct regions and suppresses noise, resulting in more accurate predictions.}
   \label{fig:onecol}
\end{figure}

Vision-Language Models (VLMs)~\cite{chen2023vlp,radford2021learning,zhang2024vision}, pretrained on large-scale image–text pairs, have achieved remarkable zero-shot generalization across a wide range of downstream tasks. Among them, CLIP~\cite{radford2021learning} has emerged as a milestone model that aligns visual and textual representations through contrastive learning, enabling powerful cross-modal understanding. Owing to its transparent architecture and strong transferability, CLIP has become a cornerstone of modern multimodal research, driving progress in computer vision~\cite{jia2021scaling,radford2021learning}, medical imaging~\cite{huang2023visual,wang2022medclip}, and robotics~\cite{ahn2022can,shridhar2022cliport}.
Despite its impressive generalization, CLIP is notably fragile under adversarial perturbations, which can severely degrade performance~\cite{kong2024patch,mao2023understanding,zhao2023evaluating,szegedy2014intriguing,fang2024clip}. Retraining such large-scale models from scratch is computationally prohibitive, leading most users to rely on public pretrained checkpoints and further amplifying the risks posed by adversarial attacks.

\begin{figure*}[!t]
  \centering
  % --- 子图 1：ViT-B/32 ---
  \begin{subfigure}[b]{0.31\linewidth}
    \centering
    \includegraphics[width=\linewidth]{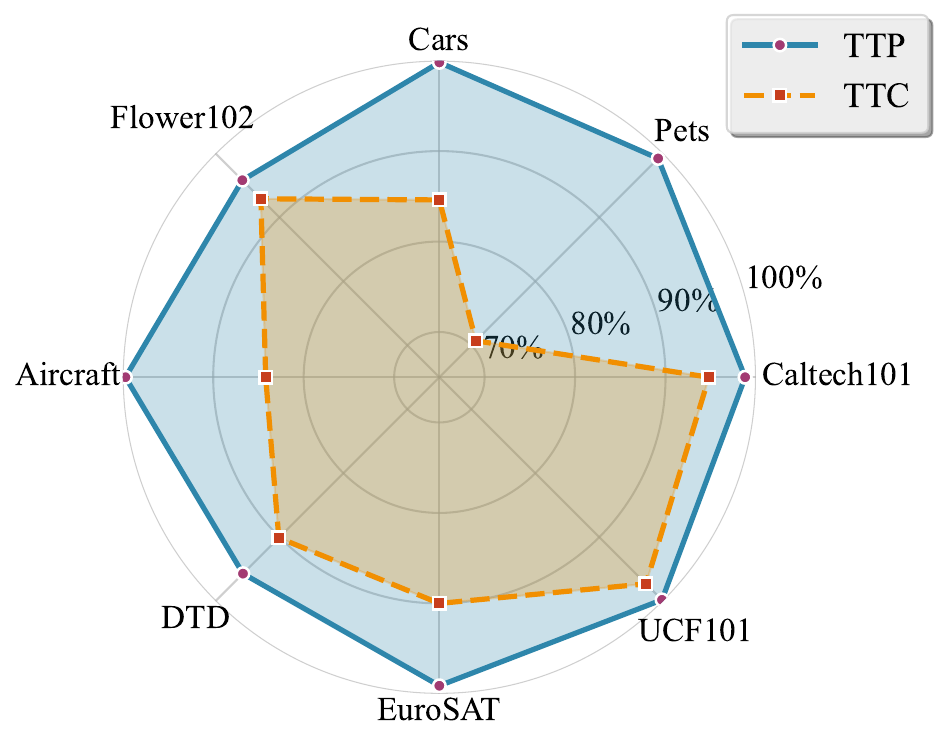}
    \caption{ViT-B/32}
    \label{fig:vsttc32}
  \end{subfigure}
  \hfill
  % --- 子图 2：ViT-B/16 ---
  \begin{subfigure}[b]{0.31\linewidth}
    \centering
    \includegraphics[width=\linewidth]{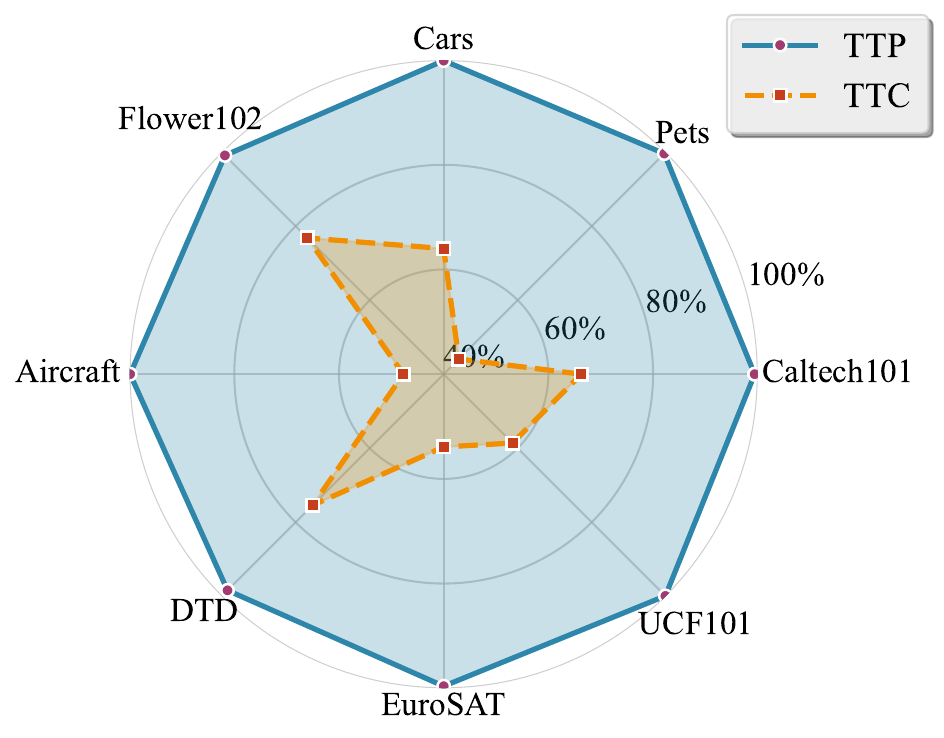}
    \caption{ViT-B/16}
    \label{fig:vsttc16}
  \end{subfigure}
  \hfill
  % --- 子图 3：ViT-L/14 ---
  \begin{subfigure}[b]{0.31\linewidth}
    \centering
    \includegraphics[width=\linewidth]{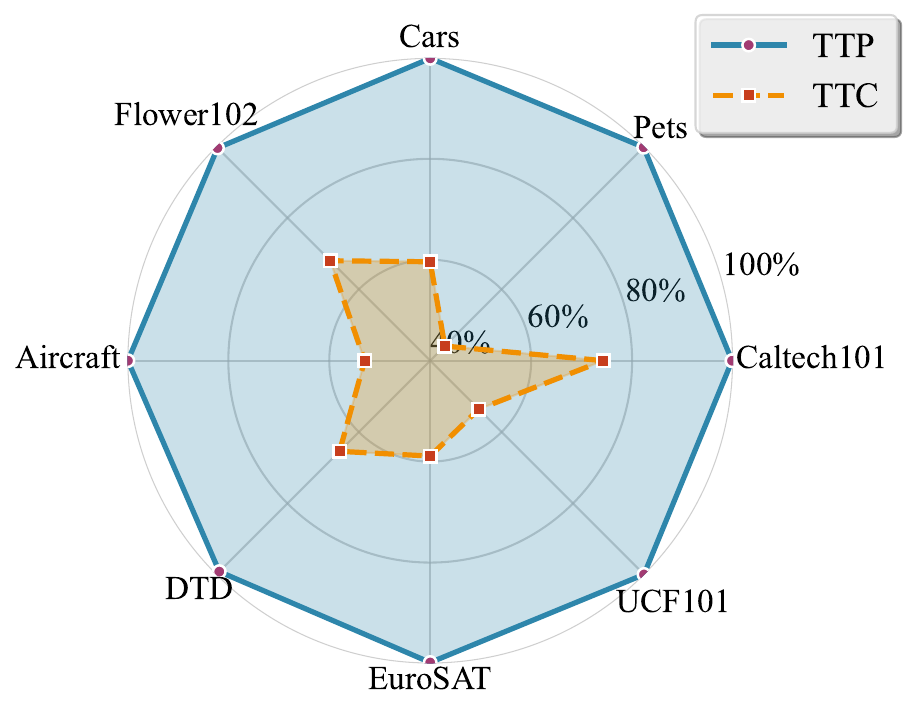}
    \caption{ViT-L/14}
    \label{fig:vsttc14}
  \end{subfigure}
  \caption{
    Detection accuracy of TTP (ours) and TTC~\cite{xing2025clip} across fine-grained classification datasets under three CLIP backbones (ViT-B/32, ViT-B/16, and ViT-L/14). All experiments are performed under the same attack strength of $\epsilon=4.0$. TTC adopts its default $L_2$-distance threshold $\tau=0.2$, but exhibits pronounced fluctuations in detection performance across both datasets and backbones, indicating its sensitivity to domain and model variations. 
    In contrast, our TTP employs a unified cosine similarity threshold $\tau=0.8$, yet maintains consistently superior detection accuracy across all settings. This demonstrates that TTP achieves not only outstanding adversarial recognition capability but also remarkable cross-dataset and cross-backbone stability, effectively mitigating the instability observed in TTC under identical conditions.
  }
  \label{fig:vsttc}
\end{figure*}

To improve robustness, prior works have explored adversarial prompt tuning (APT)~\cite{li2024one,wang2024advqdet,zhou2024few}, which learns robust text prompts via adversarial training. However, these methods require labeled adversarial data and costly retraining, and their robustness fails to generalize beyond seen categories. In contrast, test-time defense methods~\cite{sheng2025r,wang2025tapt,tong2025zero} adapt models on the fly but typically apply uniform adaptation to all inputs, leading to suboptimal performance on both robustness and clean accuracy. The recent Test-Time Counterattack (TTC)~\cite{xing2025clip} reports that adversarial examples exhibit greater stability under small perturbations, enabling discrimination between clean and adversarial inputs by measuring feature stability under slight noise. However, TTC exhibits low detection accuracy and poor generalization across datasets and model architectures (Fig.~\ref{fig:vsttc}), which severely constrains its practical utility in unseen scenarios.
% However, TTC exhibits poor generalizability across datasets and model architectures(Fig.~\ref{fig:vsttc}), which severely limits its practical utility in unseen scenarios.
% The recent Test-Time Counterattack (TTC)~\cite{xing2025clip} detects adversarial examples by leveraging perturbation stability in the feature space, yet its reliance on dataset-specific thresholds makes it unstable across datasets and backbones (see Fig.~\ref{fig:vsttc}). 

% Moreover, TTC applies perturbations indiscriminately to all samples, thereby harming clean accuracy.

To address these limitations, we propose Test-Time Padding (TTP), a simple yet highly effective framework for robust inference with CLIP. Our key insight is that image padding can restore attention disrupted by adversarial perturbations (as shown in Fig.~\ref{fig:onecol}), producing distinct similarity shifts between clean and adversarial examples. Specifically, by comparing visual embeddings before and after padding, we observe that clean samples exhibit minimal feature change, whereas adversarial examples show significant shifts. This observation enables a unified, dataset-agnostic criterion for reliable adversarial detection. As shown in Fig.~\ref{fig:vsttc}, compared to TTC, our proposed method achieves uniformly superior performance across diverse datasets and model architectures using a single, unified cosine-similarity threshold.

%%%%%%%%%
% Based on accurate detection, TTP dynamically differentiates inference strategies for clean and adversarial inputs. Clean samples are directly preserved to maintain semantic integrity and clean accuracy, while adversarial examples are further processed through a \textbf{trainable test-time padding}. 
Based on accurate detection, TTP dynamically differentiates inference strategies for clean and adversarial inputs. Clean samples are directly preserved to maintain semantic integrity and clean accuracy. For adversarial examples, our prior findings have shown that random padding can effectively recover attention patterns disrupted by adversarial perturbations. To further improve defense performance, we introduce trainable test-time padding.
During inference, multiple augmented views of each adversarial instance are generated, and padding parameters are optimized by a single step via entropy minimization over high-confidence samples. To achieve a more robust final prediction, we introduce a similarity-aware ensemble that assigns adaptive weights to each augmented view by measuring its similarity to the adversarial example’s embeddings before and after padding, thereby further strengthening the defense performance.
% To ensure stability and semantic consistency, we introduce a similarity-aware ensemble that weights each view based on its cosine similarity shift, achieving a balanced trade-off between noise suppression and feature preservation.
In summary, our main contributions are as follows:

\noindent\textbullet\ \hangindent=0.8em We show that spatial padding restores attention patterns disrupted by adversarial perturbations and use similarity shift to build a unified detector that generalizes across datasets and architectures.
% We discover that spatial padding effectively restores model attention patterns disrupted by adversarial perturbations. Leveraging this insight, we calculate embedding similarity before and after padding to devise a unified adversarial detector that generalizes across datasets and model architectures. 
% We introduce TTP, a lightweight and unified test-time defense framework that distinguishes clean and adversarial examples via a simple similarity-based criterion, enabling accurate detection without retraining.

\noindent\textbullet\ \hangindent=0.8em For detected adversarial inputs, we introduce single-step trainable padding at inference and a similarity-aware ensemble, yielding more reliable predictions.
% For detected adversarial samples, we employ a single-step trainable padding during inference to further recover model attention, and introduce a similarity-aware ensemble strategy to obtain more reliable final predictions.
% We develop a test-time padding mechanism that dynamically adapts adversarial examples during inference using entropy minimization and a similarity-aware ensemble, significantly enhancing robustness while preserving semantics.

\noindent\textbullet\ \hangindent=0.8em Combining detection with robust adaptation, we propose TTP, a two-stage detect-then-adapt defense that consistently improves robustness without sacrificing clean accuracy across multiple CLIP backbones and fine-grained datasets, significantly surpassing prior test-time defenses.

\section{Related Work}

\subsection{Adversarial Attacks and Defenses}
Adversarial attacks expose the vulnerability of neural networks by introducing imperceptible perturbations that mislead predictions~\cite{goodfellow2014explaining,szegedy2014intriguing,madry2017towards}. Gradient-based white-box attacks, such as FGSM~\cite{goodfellow2014explaining} and BIM~\cite{kurakin2018adversarial}, generate adversarial examples through iterative or single-step gradient updates~\cite{ma2024imbalanced, zhang2022towards, zhou2023advclip}. Among them, PGD (Projected Gradient Descent)~\cite{madry2017towards} has become the \textit{de facto} standard benchmark for evaluating model robustness due to its strong attack capability and generality. In contrast, black-box attacks, including transfer-based~\cite{lu2023set} and query-based variants~\cite{fang2024one,he2023sa,zhang2022towards,zhang2024universal,wang2024transferable}, operate without gradient access, reflecting more realistic threat scenarios. These attack families collectively demonstrate the fragility of high-level representations even in large-scale pretrained VLMs such as CLIP, motivating the study of efficient and model-agnostic defenses.

\indent Existing defenses focus primarily on improving the stability of the model against perturbations~\cite{madry2017towards,liang2025comprehensive,wu2020adversarial,zhang2019theoretically}. Adversarial training and its variants remain the most effective but require extensive labeled data and retraining. For vision-language models, robust fine-tuning strategies such as TeCoA~\cite{mao2023understanding} and APT~\cite{li2024one}, which adjust prompts or image encoders through adversarially augmented data, achieving improved robustness but at high computational and data costs. Our work follows an alternative direction—\textbf{test-time robust adaptation}, which optimizes model behavior on-the-fly without modifying pretrained weights.

\subsection{Test-Time Adaptation and Defense for VLMs}
Test-time adaptation (TTA) has emerged as an effective strategy to improve generalization on unseen domains during inference~\cite{karmanov2024efficient, yu2023benchmarking,sui2025just, meng2025black, ma2023swapprompt}. Representative approaches such as TENT~\cite{wang2021tent}, TPT~\cite{shu2022test}, and DiffTPT~\cite{feng2023diverse} adapt model parameters or prompts by minimizing prediction entropy across augmented test views. Ensemble-based methods~\cite{abdul2023align,zanella2024test} further enhance stability by aggregating multiple views or prompt variants. These methods effectively mitigate distribution shifts and improve zero-shot accuracy but overlook robustness concerns, as they assume clean input data and stationary test distributions.

\indent Recent studies extend TTA to adversarial scenarios, with the aim of improving the robustness of the model without retraining. TAPT~\cite{wang2025tapt} optimizes both textual and visual prompts by minimizing a marginal entropy objective, while R-TPT~\cite{sheng2025r} takes a different route, decomposing the entropy objective into a pointwise loss to adapt only the text prompt during inference. Although both approaches achieve notable robustness gains, they share a fundamental flaw: they apply a uniform adaptation strategy to all inputs. This design is inherently suboptimal, as the conflicting adaptation objectives for clean and adversarial inputs limit the overall performance. 

TTC~\cite{xing2025clip} employs a different two-stage paradigm. It first distinguishes between inputs by measuring their feature embedding shifts under small perturbations, enabling separate optimization. However, its detection mechanism suffers from poor accuracy and limited generalization. In contrast, our Test-Time Padding (TTP) improves this two-stage approach. By enabling highly accurate and generalizable adversarial detection, it applies a dedicated adaptation mechanism only where needed, substantially enhancing robustness without compromising the integrity of clean data.

% Methods such as R-TPT~\cite{sheng2025r} and TAPT~\cite{wang2025tapt} enhance adversarial robustness by adapting prompts on adversarial test samples, demonstrating that test-time optimization can effectively mitigate attack-induced degradation. 

% However, they adopt identical adaptation strategies for both clean and adversarial inputs, which limits the optimization space since the two exhibit distinct adaptation objectives. 

% In contrast, our \textbf{Test-Time Padding (TTP)} detects adversarial instances and applies a dedicated adaptation mechanism.

\begin{figure*}[!t]
  \centering
  % \fbox{\rule{0pt}{2in} \rule{0.95\linewidth}{0pt}}
   \includegraphics[width=0.95\linewidth]{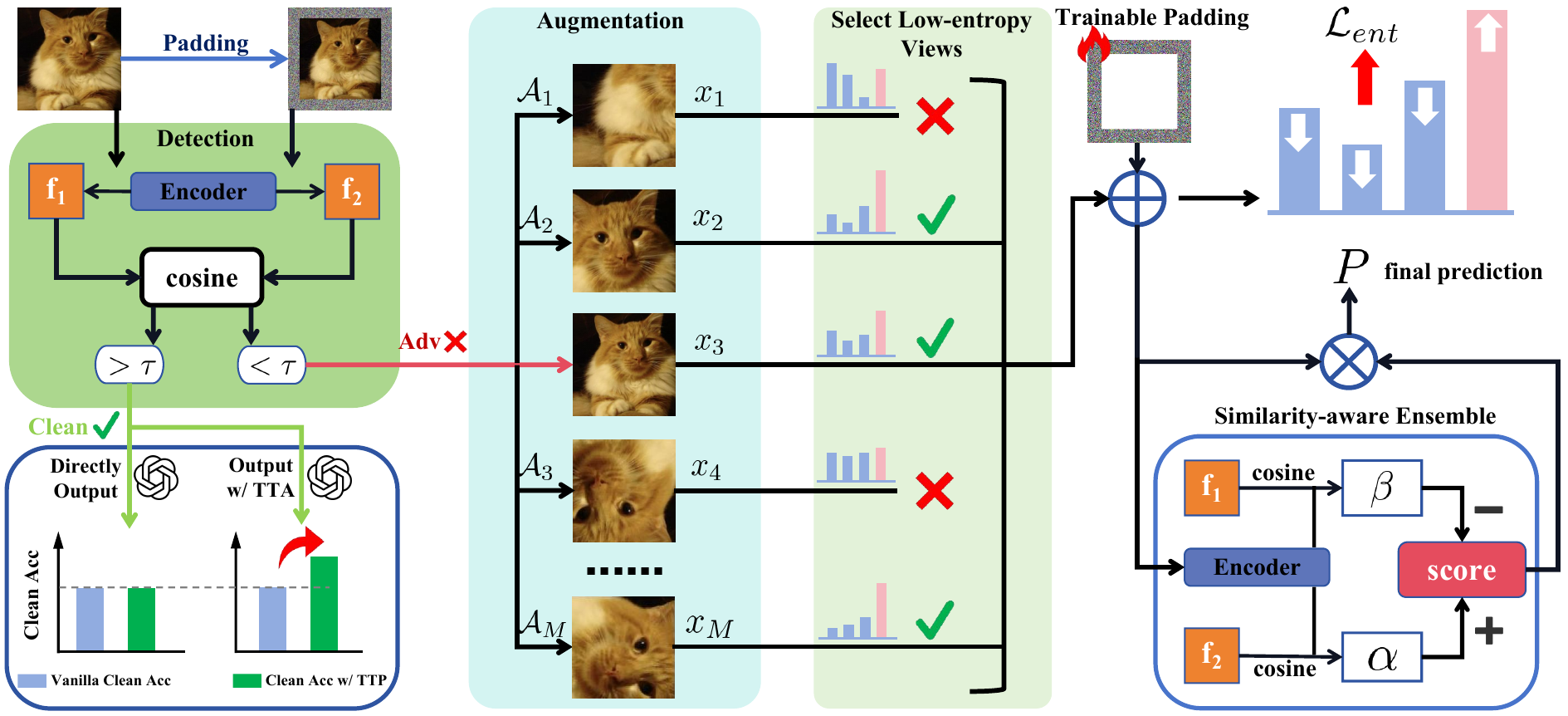}
    \caption{Overview of the proposed \textbf{Test-Time Padding (TTP)} pipeline. Given an input sample, CLIP image encoder features are extracted before and after applying padding. Their cosine similarity difference is compared with a universal threshold to distinguish clean versus adversarial inputs. Clean samples are directly recognized without adaptation. For adversarial examples, \textbf{trainable test-time padding} is activated to optimize padding parameters by entropy minimization using augmented views with low entropy. A \textbf{similarity-aware ensemble} then aggregates predictions across selected high-confidence views, ensuring a more reliable final prediction. Together, TTP enables accurate adversarial detection and adaptation-driven robustness improvement.
}
   \label{fig:Mechanism}
\end{figure*}

\section{Methodology}

\subsection{Preliminaries}

CLIP (Contrastive Language–Image Pre-training) is a widely adopted vision–language model with a dual-tower architecture, consisting of an image encoder $F(\cdot)$ and a text encoder $G(\cdot)$. It is pre-trained using a large-scale set of image–text pairs through a contrastive learning objective that aligns visual and textual representations in a shared embedding space. Specifically, for each image–text pair $(x, t)$, CLIP maximizes the cosine similarity between the corresponding encoded features while minimizing similarities with mismatched pairs. This pre-training paradigm enables CLIP to acquire rich multimodal knowledge and strong zero-shot generalization capabilities across diverse downstream tasks.

Given a $C$-way classification problem with class names $\{t_c\}_{c=1}^{C}$, CLIP constructs textual features by feeding a hand-crafted prompt template (e.g., “\emph{a photo of a [CLASS]}”) into the text encoder:
\begin{equation}
g_c = G(\mathrm{prompt}(t_c)), \quad c = 1, \dots, C,
\end{equation}
where $g_c \in \mathbb{R}^d$ denotes the textual feature of class $c$. For a test image $x_i$, the image encoder $F(\cdot)$ extracts its feature representation:
\begin{equation}
f_i = F(x_i).
\end{equation}
where $f_i \in \mathbb{R}^d$.
CLIP then computes the cosine similarity between $f_i$ and each $g_c$, and applies a softmax operation to obtain the probability that $x_i$ belongs to class $c$:
\begin{equation}
p_c(x_i) = \frac{\exp(\cos(f_i, g_c)/\tau)}{\sum_{j=1}^{C} \exp(\cos(f_i, g_j)/\tau)},
\label{eq:clipprob}
\end{equation}
where $\cos(\cdot)$ denotes the cosine similarity and $\tau$ is the temperature hyperparameter, typically set to 0.01. Benefiting from large-scale multimodal pretraining, CLIP exhibits excellent zero-shot recognition and strong cross-domain generalization.

\subsection{Test-Time Padding (TTP)}
\label{sec:tttp}
As illustrated in Fig.~\ref{fig:Mechanism}, we propose \textbf{Test-Time Padding (TTP)}, a lightweight test-time defense for CLIP that operates directly in the input space. Our approach is motivated by the key observation that applying padding to adversarial inputs partially restores disrupted attention, yielding a characteristic cosine-similarity shift between embeddings before and after padding. Clean samples exhibit minimal shift, whereas adversarial examples show pronounced change. This “similarity shift” provides a unified, model- and dataset-agnostic criterion for reliable detection without retraining.

Building on this, TTP forms a coherent three-stage pipeline: (1) we first perform \textbf{adversarial example detection} using the similarity shift to separate clean and adversarial inputs, allowing clean samples to bypass adaptation and thus preserving clean accuracy. (2) For detected adversarial examples, we employ \textbf{trainable test-time padding}, where instance-specific padding parameters are optimized by minimizing prediction entropy across confident augmented views, thereby better restoring the model's attention. (3) Finally, a \textbf{similarity-aware ensemble} assigns distinct weights to each augmented view, leading to more robust predictions.

\noindent\textbf{Adversarial example detection.}  
Given a test sample $x$, we first apply a fixed padding operation $P^\mathrm{fix}(x)$ and encode both $x$ and $P^\mathrm{fix}(x)$ using the frozen CLIP image encoder:
\begin{equation}
z = F(x), \quad z^{\text{pad}} = F(P^\mathrm{fix}(x)).
\end{equation}
The cosine similarity
\begin{equation}
s = \frac{z \cdot z^{\text{pad}}}{\|z\|\|z^{\text{pad}}\|}
\label{eq6}
\end{equation}
is then compared against a threshold $\tau$.  
If $s > \tau$, the sample is considered clean and is classified directly. Otherwise, it is treated as adversarial and passed on to the subsequent adaptive stages.  
This step addresses a critical limitation of prior test-time defense methods, which apply the same adaptation strategy to all inputs, by enabling sample-specific adaptation based on reliable detection.  
High detection accuracy also ensures that clean samples are largely unaffected, which preserves clean accuracy and enables the combination with other TTA methods to further improve performance.

\noindent\textbf{Trainable test-time padding.}  
For detected adversarial examples, following the widely adopted marginal entropy minimization paradigm in test-time adaptation~\cite{sheng2025r, wang2025tapt, shu2022test}, we generate multiple augmented views $\{x_i\}_{i=1}^N$ using stochastic transformations (e.g., random resize, crop, color jitter). Each view is encoded and classified to obtain prediction probabilities $p_c(x_i)$ and their Shannon entropies:
\begin{equation}
H_i = -\sum_{c=1}^C p_c(x_i) \log p_c(x_i).
\end{equation}
The top-K low-entropy views form the confident subset $B$ used for subsequent adaptation.  Unlike prior test-time defenses that adapt the textual prompt, our proposed method optimizes instance-specific image padding parameters at inference, for restoring attention patterns disrupted by adversarial perturbations. More specifically, a lightweight trainable padding module $P_\theta(\cdot)$ is firstly applied to $B$ to obtain the corresponding augmented entropy $H_i^{pad}$. The parameter $\theta$ is then updated for a single step by minimizing the average entropy of the padded views:
\begin{equation}
\mathcal{L}_{\text{ent}} = \frac{1}{|B|}\sum_{i \in B} H_i^{\text{pad}}, \quad
\theta \leftarrow \theta - \eta \nabla_\theta \mathcal{L}_{\text{ent}}.
\label{eq8}
\end{equation}
Once updated, the padding module helps the model maintain stable predictions across multiple augmented adversarial views. Compared with random padding, this average entropy training strategy substantially reduces noise when restoring the model’s attention patterns.
% This optimization drives the trainable padding to produce outputs with higher confidence on the selected low-entropy views, encouraging the model to focus on semantically meaningful regions and retain the essential content of the image.

\noindent\textbf{Similarity-aware ensemble.}
After padding adaptation, we employ a similarity-based ensemble strategy, weighting the predictions from selected low-entropy views to obtain a more reliable final output. Specifically, for a given adversarial input $x_\mathrm{adv}$, we first extract its image embedding before and after padding using the image encoder, yielding $z_\mathrm{adv} = F(x_\mathrm{adv})$ and $z_\mathrm{adv}^\mathrm{pad} = F(P^\mathrm{fix}(x_\mathrm{adv}))$, respectively. For each selected augmented view $x_i$, we apply the trained padding module to obtain $P_\theta(x_i)$, then compute its image embedding $z_i^\mathrm{pad} = F(P_\theta(x_i))$. Finally, the weight $w_i$ assigned to each $x_i$ is calculated as follows:
\begin{align}
\alpha_i &= \cos(z_i^\mathrm{pad}, z^{\text{pad}}_{\text{adv}}),\\
\beta_i  &= \cos(z_i^\mathrm{pad}, z_{\text{adv}}),\\
s_i &= \alpha_i - \beta_i, \quad
w_i = \frac{\exp(s_i)}{\sum_{j\in B}\exp(s_j)}.
\label{eq11}
\end{align}
Here, $\alpha_i$ is the cosine similarity between the padded embedding of $x_i$ and the padded embedding of $x_\mathrm{adv}$, while $\beta_i$ measures similarity to the original adversarial embedding without padding. Because adversarial perturbations substantially distort image features, we prefer padded views that are farther from the original adversarial embedding (smaller $\beta_i$). However, simply increasing the distance from the adversarial example is not sufficient, as our objective is to achieve more accurate predictions. Our previous experiments demonstrate that trainable padding can effectively restore model attention, leading to highly similar attention maps with clean samples. Consequently, we also prefer augmented views closer to the padded adversarial embedding (larger $\alpha_i$). Combining these insights, we use $s_i = \alpha_i - \beta_i$ as the view score and apply a softmax to obtain $w_i$, yielding a similarity-aware ensemble that prioritizes the most reliable views to maximize prediction accuracy. The final prediction is then computed as follows:
\begin{equation}
p_{\text{final}} = \arg\max_c \sum_{i \in B} w_i p_c(P_\theta(x_i)).
\label{eq12}
\end{equation}

\begin{table*}[ht]
\centering
\resizebox{\linewidth}{!}{%
\begin{tabular}{l|cc|cc|cc|cc|cc|cc|cc|cc|cc}
\toprule
Method & \multicolumn{2}{c|}{Caltech101} & \multicolumn{2}{c|}{Pets} & \multicolumn{2}{c|}{Cars} & \multicolumn{2}{c|}{Flower102} & \multicolumn{2}{c|}{Aircraft} & \multicolumn{2}{c|}{DTD} & \multicolumn{2}{c|}{EuroSAT} & \multicolumn{2}{c|}{UCF101} & \multicolumn{2}{c}{Avg.} \\
& Acc. & Rob. & Acc. & Rob. & Acc. & Rob. & Acc. & Rob. & Acc. & Rob. & Acc. & Rob. & Acc. & Rob. & Acc. & Rob. & Acc. & Rob. \\
\midrule
CLIP~\cite{radford2021learning} & 91.4 & \cellcolor{gray!20}0.2 & 85.1 & \cellcolor{gray!20}0.0 & 60.1 & \cellcolor{gray!20}0.0 & 64.0 & \cellcolor{gray!20}0.0 & 18.1 & \cellcolor{gray!20}0.0 & 43.0 & \cellcolor{gray!20}0.0 & 35.8 & \cellcolor{gray!20}0.0 & 61.6 & \cellcolor{gray!20}0.0 & 57.4 & \cellcolor{gray!20}0.0 \\
TTC~\cite{xing2025clip} & 86.5 & \cellcolor{gray!20}22.7 & 83.5 & \cellcolor{gray!20}11.8 & 48.1 & \cellcolor{gray!20}2.3 & 64.3 & \cellcolor{gray!20}3.2 & 18.2 & \cellcolor{gray!20}1.0 & 37.3 & \cellcolor{gray!20}4.7 & \textbf{53.0} & \cellcolor{gray!20}3.0 & 62.6 & \cellcolor{gray!20}6.1 & 56.7 & \cellcolor{gray!20}6.8\\
Ensemble & 88.2 & \cellcolor{gray!20}74.9 & 75.0 & \cellcolor{gray!20}52.5 & 51.7 & \cellcolor{gray!20}25.9 & 58.1 & \cellcolor{gray!20}36.1 & 16.4 & \cellcolor{gray!20}7.9 & 39.8 & \cellcolor{gray!20}28.6 & 30.8 & \cellcolor{gray!20}11.9 & 54.9 & \cellcolor{gray!20}36.9 & 50.1 & \cellcolor{gray!20}34.3 \\
MTA~\cite{zanella2024test} & \textbf{92.0} & \cellcolor{gray!20}76.3 & \textbf{86.3} & \cellcolor{gray!20}53.6 & \textbf{63.4} & \cellcolor{gray!20}26.4 & \textbf{64.4} & \cellcolor{gray!20}36.5 & \textbf{20.2} & \cellcolor{gray!20}8.2 & \textbf{43.8} & \cellcolor{gray!20}28.8 & 34.6 & \cellcolor{gray!20}11.3 & \textbf{63.3} & \cellcolor{gray!20}39.1 & \textbf{58.3} & \cellcolor{gray!20}35.0 \\
R-TPT~\cite{sheng2025r} & 90.6 & \cellcolor{gray!20}76.4 & 84.5 & \cellcolor{gray!20}55.8 & 63.1 & \cellcolor{gray!20}28.4 & 62.6 & \cellcolor{gray!20}37.6 & 19.1 & \cellcolor{gray!20}9.2 & 42.1 & \cellcolor{gray!20}29.1 & 32.0 & \cellcolor{gray!20}5.1 & 62.8 & \cellcolor{gray!20}41.0 & 57.3 & \cellcolor{gray!20}35.3 \\
TTP (Ours) & 90.9 & \cellcolor{gray!20}\textcolor{cvprpink}{\textbf{81.8}} & 84.7 & \cellcolor{gray!20}\textcolor{cvprpink}{\textbf{61.0}} & 59.8 & \cellcolor{gray!20}\textcolor{cvprpink}{\textbf{29.8}} & 63.6 & \cellcolor{gray!20}\textcolor{cvprpink}{\textbf{42.0}} & 18.0 & \cellcolor{gray!20}\textcolor{cvprpink}{\textbf{10.3}} & 42.8 & \cellcolor{gray!20}\textcolor{cvprpink}{\textbf{32.2}} & 35.6 & \cellcolor{gray!20}\textcolor{cvprpink}{\textbf{14.1}} & 61.3 & \cellcolor{gray!20}\textcolor{cvprpink}{\textbf{46.6}} & 57.1 & \cellcolor{gray!20}\textcolor{cvprpink}{\textbf{39.7}} \\
\bottomrule
\end{tabular}%
}
\caption{Clean (Acc.) and adversarial (Rob.) accuracy (\%) on \textbf{fine-grained classification datasets} with pre-trained CLIP-ViT-B/32 ($\epsilon=4.0$). The best results of clean accuracy are \textbf{bolded}, and the best results of robustness are \textbf{\textcolor{cvprpink}{bolded}}.}
\label{tab:vitb32}
\end{table*}

\begin{table*}[h]
\centering
\resizebox{\linewidth}{!}{%
\begin{tabular}{l|cc|cc|cc|cc|cc|cc|cc|cc|cc}
\toprule
Method & \multicolumn{2}{c|}{Caltech101} & \multicolumn{2}{c|}{Pets} & \multicolumn{2}{c|}{Cars} & \multicolumn{2}{c|}{Flower102} & \multicolumn{2}{c|}{Aircraft} & \multicolumn{2}{c|}{DTD} & \multicolumn{2}{c|}{EuroSAT} & \multicolumn{2}{c}{UCF101} & \multicolumn{2}{c}{Avg.} \\
& Acc. & Rob. & Acc. & Rob. & Acc. & Rob. & Acc. & Rob. & Acc. & Rob. & Acc. & Rob. & Acc. & Rob. & Acc. & Rob. & Acc. & Rob. \\
\midrule
CLIP~\cite{radford2021learning} & 94.0 & \g0.0 & \textbf{88.3} & \g0.0 & 65.5 & \g0.0 & 67.4 & \g0.0 & 23.9 & \g0.0 & 44.4 & \g0.0 & 42.2 & \g0.0 & 65.2 & \g0.0 & 61.4 & \g0.0 \\
TTC~\cite{xing2025clip} & 87.6 & \g8.4 & 82.3 & \g10.4 & 55.0 & \g2.9 & \textbf{69.0} & \g7.4 & 23.3 & \g0.5 & 41.0 & \g4.5 & \textbf{47.4} & \g0.4 & 65.8 & \g1.6 & 58.9 & \g4.5\\
Ensemble & 91.9 & \g74.7 & 86.2 & \g51.2 & 65.7 & \g26.0 & 65.9 & \g36.3 & 23.4 & \g8.7 & 43.2 & \g25.1 & 28.2 & \g2.2 & 63.0 & \g30.6 & 58.4 & \g31.8 \\
% C-TPT~\cite{wang2023improving} & 93.9 & 0.0 & 88.2 & 0.0 & 65.8 & 0.0 & \textbf{69.6} & 0.0 & 23.9 & 0.0 & 45.9 & 0.0 & 42.3 & 0.0 & 65.5 & 0.0 & 61.9 & 0.0 \\
MTA~\cite{zanella2024test} & \textbf{94.3} & \g72.1 & 88.0 & \g51.8 & \textbf{67.7} & \g18.5 & 67.4 & \g27.9 & \textbf{25.0} & \g4.3 & \textbf{46.5} & \g16.2 & 42.5 & \g1.2 & \textbf{67.5} & \g27.5 & \textbf{62.3} & \g27.4 \\
R-TPT~\cite{sheng2025r} & 93.7 & \g82.0 & 87.2 & \g60.2 & 67.0 & \g34.7 & 68.7 & \g44.6 & 23.9 & \g13.2 & 46.4 & \g32.8 & 34.7 & \g8.5 & 67.2 & \g43.2 & 61.1 & \g39.9 \\
TTP (Ours) & 93.5 & \g\textcolor{cvprpink}{\textbf{82.3}} & \textbf{88.3} & \g\textcolor{cvprpink}{\textbf{64.7}} & 65.4 & \g\textcolor{cvprpink}{\textbf{37.4}} & 67.3 & \g\textcolor{cvprpink}{\textbf{47.2}} & 23.9 & \g\textcolor{cvprpink}{\textbf{14.8}} & 44.1 & \g\textcolor{cvprpink}{\textbf{36.0}} & 42.0 & \g\textcolor{cvprpink}{\textbf{14.5}} & 65.0 & \g\textcolor{cvprpink}{\textbf{47.2}} & 61.2 & \g\textcolor{cvprpink}{\textbf{42.9}} \\
\bottomrule
\end{tabular}%
}
\caption{Adversarial (Rob.) and Clean (Acc.) accuracy (\%) on \textbf{fine-grained classification datasets} with pre-trained CLIP-ViT-B/16 ($\epsilon=4.0$). The best results of clean accuracy are \textbf{bolded}, and the best results of robustness are \textbf{\textcolor{cvprpink}{bolded}}.}
\label{tab:vitb16}
\end{table*}

\begin{table*}[h]
\centering
\resizebox{\linewidth}{!}{% 可选：自动适应行宽，可删除
\begin{tabular}{l|cc|cc|cc|cc|cc|cc|cc|cc|cc}
\toprule
Method & \multicolumn{2}{c|}{Caltech101} & \multicolumn{2}{c|}{Pets} & \multicolumn{2}{c|}{Cars} & \multicolumn{2}{c|}{Flower102} & \multicolumn{2}{c|}{Aircraft} & \multicolumn{2}{c|}{DTD} & \multicolumn{2}{c|}{EuroSAT} & \multicolumn{2}{c}{UCF101} & \multicolumn{2}{c}{Avg.} \\

& Acc. & Rob. & Acc. & Rob. & Acc. & Rob. & Acc. & Rob. & Acc. & Rob. & Acc. & Rob. & Acc. & Rob. & Acc. & Rob. & Acc. & Rob. \\
\midrule
CLIP~\cite{radford2021learning} & 95.2 & \g0.1 & 93.1 & \g0.0 & 76.8 & \g0.0 & 76.2 & \g0.0 & 30.0 & \g0.0 & 52.4 & \g0.0 & 55.1 & \g0.0 & 73.7 & \g0.0 & \textbf{69.1} & \g0.0 \\
TTC~\cite{xing2025clip} & 88.7 & \g7.7 & 92.2 & \g7.6 & 67.8 & \g2.2 & \textbf{76.5} & \g7.5 & 31.7 & \g0.5 & 49.7 & \g6.2 & \textbf{64.1} & \g0.2 & \textbf{75.0} & \g2.2 & 68.2 & \g4.3\\
Ensemble & 94.9 & \g83.6 & 93.4 & \g63.5 & 76.3 & \g40.5 & 75.0 & \g48.6 & 31.7 & \g12.7 & 51.3 & \g31.3 & 38.7 & \g11.1 & 71.7 & \g48.3 & 66.6 & \g42.5 \\
% TPT~\cite{shu2022test} & \textbf{95.9} & 0.2 & 93.8 & 0.0 & 78.0 & 0.0 & 76.9 & 0.0 & 31.6 & 0.0 & \textbf{55.1} & 0.0 & 51.8 & 0.0 & \textbf{74.7} & 0.0 & 69.7 & 0.0 \\
% C-TPT~\cite{wang2023improving} & 95.6 & 0.1 & \textbf{94.3} & 0.0 & 77.4 & 0.0 & 76.3 & 0.0 & 30.4 & 0.0 & \textbf{55.4} & 0.0 & 54.0 & 0.0 & 75.1 & 0.0 & \textbf{69.8} & 0.0 \\
MTA~\cite{zanella2024test} & \textbf{95.8} & \g83.1 & \textbf{93.7} & \g64.9 & \textbf{78.4} & \g36.6 & 76.1 & \g44.2 & \textbf{32.7} & \g8.0 & 53.4 & \g27.2 & 47.8 & \g7.5 & 74.7 & \g47.5 & \textbf{69.1} & \g39.9 \\
R-TPT~\cite{sheng2025r} & 95.7 & \g88.2 & \textbf{93.7} & \g72.9 & 77.2 & \g49.1 & 76.2 & \g55.6 & 31.7 & \g17.2 & \textbf{54.0} & \g38.0 & 44.3 & \g20.4 & 74.3 & \g55.6 & 68.4 & \g49.6 \\
TTP (Ours) & 95.1 & \g\textcolor{cvprpink}{\textbf{88.6}} & 93.1 & \g\textcolor{cvprpink}{\textbf{76.3}} & 76.8 & \g\textcolor{cvprpink}{\textbf{51.1}} & 76.1 & \g\textcolor{cvprpink}{\textbf{58.7}} & 29.2 & \g\textcolor{cvprpink}{\textbf{17.7}} & 52.3 & \g\textcolor{cvprpink}{\textbf{41.3}} & 55.0 & \g\textcolor{cvprpink}{\textbf{21.6}} & 73.6 & \g\textcolor{cvprpink}{\textbf{57.4}} & 68.9 & \g\textcolor{cvprpink}{\textbf{51.6}} \\
\bottomrule
\end{tabular}
}
\caption{Adversarial (Rob.) and Clean (Acc.) accuracy (\%) on \textbf{fine-grained classification datasets} with pre-trained CLIP-ViT-L/14 ($\epsilon=4.0$). The best results of clean accuracy are \textbf{bolded}, and the best results of robustness are \textbf{\textcolor{cvprpink}{bolded}}.}
\label{tab:vitl14}
\end{table*}

\begin{table*}[ht]
\centering
\resizebox{0.8\textwidth}{!}{
\begin{tabular}{l|cccccccc|c}
\toprule
Method & Caltech101 & Pets & Cars & Flower102 & Aircraft & DTD & EuroSAT & UCF101 & Avg. \\
\midrule
TTP (0)           & 98.8 & 99.2 & \textbf{99.8} & 95.8 & 99.7 & 95.7 & \textbf{99.1} & \textbf{99.8} & 98.5 \\
TTP (random)              & 97.5 & 98.9 & 97.6 & 96.9 & 92.3 & 90.7 & 94.9 & 95.5 & 95.8 \\
TTP (255)         & \textbf{98.9} & \textbf{99.4} & 99.6 & \textbf{97.1} & \textbf{99.8} & \textbf{96.2} & 98.5 & 99.7 & \textbf{98.7} \\
% TTC~\cite{xing2025clip}     & 94.8 & 70.7 & 84.6 & 92.9 & 84.2 & 90.1 & 90.0 & 97.3 & 88.7 \\
\bottomrule
\end{tabular}}
\caption{Detection accuracy (\%) of TTP using different padding patterns on \textbf{fine-grained classification datasets} with pre-trained CLIP-ViT-B/32 ($\epsilon=4.0$). The best results are \textbf{(bolded)}.}
% TTP results are under padding size 32 (different padding types). 
\label{tab:padding_type_meanacc}
\end{table*}

By integrating adversarial detection, instance-specific padding adaptation, and similarity-aware ensemble, TTP provides a novel framework for test-time robustness enhancement. Operating entirely in the padding space, it requires neither text prompt tuning nor any model architectural knowledge or model modifications, making it a lightweight, efficient, and transferable defense mechanism. The total process of TTP is outlined in Algorithm~\ref{alg:ttp}.

\begin{algorithm}[t]
\caption{Test-Time Padding (TTP)}
\label{alg:ttp}
\small
\KwIn{Test sample $x$, frozen CLIP predictor $p_c(\cdot)$, detection threshold $\tau$,
number of augmented views $N$, fixed padding $P^{\mathrm{fix}}(\cdot)$ and trainable padding $P_\theta(\cdot)$.}
\KwOut{Robust prediction $p_{\mathrm{final}}$}

Encode $x$ and $P^{\mathrm{fix}}(x)$ to measure similarity $s$ (Eq.~\ref{eq6})\;
\If{$s > \tau$}{
    \Return{$p_c(x)$}\;
}
\Else{ % adversarial case
    Generate $N$ augmented views $\{x_i\}_{i=1}^N$ \;
    Select low-entropy subset $B$\;
    Update $P_\theta(\cdot)$ by minimizing average entropy (Eq.~\ref{eq8})\;
    Compute weight $w_i$ of each $x_i$ (Eq.~\ref{eq11})\;
    Obtain final weighted prediction:
    $p_{\mathrm{final}} = \arg \max_c \sum_{i \in B} w_i \, p_c(P_\theta(x_i))$ (Eq.~\ref{eq12})\;
}
\Return{$p_{\mathrm{final}}$}\;
\end{algorithm}

\section{Experiments}

\subsection{Setup}

\noindent \textbf{Datasets and Models.} To evaluate the effectiveness of our test-time defense for vision-language models (VLMs), we conduct experiments on eight fine-grained classification datasets. These datasets cover a diverse range of domains, including general objects (Caltech101~\cite{fei2004learning}), animals (OxfordPets~\cite{parkhi2012cats}), plants (Flower102~\cite{nilsback2008automated}), vehicles (Cars~\cite{krause20133d}, Aircraft~\cite{maji2013fine}), textures (DTD~\cite{cimpoi2014describing}), satellite imagery (EuroSAT~\cite{helber2019eurosat}), and video-based actions (UCF101~\cite{soomro2012dataset}). Our experiments focus on CLIP as the backbone model, with three architectures considered: ViT-B/32, ViT-B/16 and ViT-L/14. 

% To further demonstrate the generalization capability of our proposed TTP, we conduct evaluations on CLIP-ResNet50 using the \textbf{ImageNet}\cite{deng2009imagenet} dataset and four ImageNet out-of-distribution (OOD) benchmarks besides fine-grained classification datasets. Specifically, \textbf{ImageNet-A}\cite{hendrycks2021natural} includes 200 classes and 7,500 natural adversarial examples collected via an adversarial filtration process. \textbf{ImageNet-V2}\cite{recht2019imagenet} contains 10,000 natural images covering the same 1,000 categories as ImageNet but sourced independently. \textbf{ImageNet-R}\cite{hendrycks2021many} provides 30,000 images rendered in diverse artistic and texture styles, resulting in notable domain differences from the original ImageNet. \textbf{ImageNet-S}\cite{wang2019learning} comprises 50,889 sketch-style images that share the same label space with ImageNet. Since TTP is designed to defend against test-time adversarial perturbations, our method does not require access to the training sets of any of these datasets.

\indent \textbf{Evaluation and Baselines.} To assess adversarial robustness, we follow the evaluation protocol of R-TPT~\cite{sheng2025r}, reporting accuracy on adversarial examples generated by PGD. We compare our method against two test-time defense baselines for CLIP: TTC~\cite{xing2025clip} and the state-of-the-art method R-TPT~\cite{sheng2025r}, as well as two test-time adaptation baselines: Ensemble and MTA~\cite{zanella2024test}, and the zero-shot predictions from the original CLIP model. Here, Ensemble denotes simple averaging of predictions across all augmented views. All methods, including ours, use only the CLIP backbone and \textit{AugMix} augmentation~\cite{hendrycksaugmix}, without incorporating additional foundation models (e.g., LLMs) or external knowledge.

\indent \textbf{Implementation Details.} For adversarial evaluation, we employ the PGD attack with 100 iterations and a perturbation bound of $\epsilon = 4.0$. 
The trainable padding module uses a padding size of 32, and the adversarial detection threshold is set to 0.8. 
During test-time adaptation, the padding parameters are randomly initialized within $[0,10]$ (corresponding to pixel values in $[0,255]$) and optimized with a single update step. 
The learning rate for padding is set to 5, and the augmented batch size is 64. All experiments are conducted on NVIDIA RTX 3090 GPUs.

\subsection{Results}
\label{4_2}
\noindent

\noindent \textbf{Results on various datasets.} We first evaluate TTP on fine-grained classification datasets. As shown in Tab.~\ref{tab:vitb32}, with ViT-B/32 as the backbone, TTP achieves an average adversarial accuracy of 39.7\% under 100-step PGD attacks with $\epsilon=4/255$. 
Across all benchmark datasets and baselines
% (i.e.,  TTC~\cite{xing2025clip}, Ensemble, MTA~\cite{zanella2024test}, and R-TPT~\cite{sheng2025r})
,  TTP consistently delivers superior adversarial robustness, demonstrating that trainable test-time padding significantly improves model resilience against adversarial perturbations. Compared to the state-of-the-art method R-TPT, TTP achieves an average performance improvement of 4.4\%. 

Notably, TTC, which also adopts a detection-then-defense strategy, exhibits poor robustness under such a strong attack, with an average adversarial accuracy of just 6.8\%. This is mainly due to its low detection accuracy on adversarial examples in high-strength attack scenarios. As illustrated earlier in Fig.~\ref{fig:vsttc}, TTC suffers from limited detection accuracy across different datasets and model architectures. If the detection step fails to differentiate clean samples from adversarial ones, adversarial examples may be misclassified as clean and evade robust adaptation, while clean samples might erroneously undergo robust adaptation, resulting in a loss of clean accuracy. In contrast, TTP achieves nearly 100\% detection accuracy across diverse model architectures and datasets (also shown in Fig.~\ref{fig:vsttc}), greatly improving the targeting of subsequent robust adaptation and substantially enhancing adversarial robustness.

\textbf{Results of different CLIP backbones.} For experimental completeness, we next explore the transferability of TTP by applying it to other commonly used CLIP backbones, including CLIP-ViT-B/16 and CLIP-ViT-L/14. As shown in Tab.~\ref{tab:vitb16} and \ref{tab:vitl14}, TTP consistently enhances robustness across all benchmarks relative to baseline methods. This robustness gain is maintained even as the model scale increases significantly, indicating that TTP generalizes well to large-scale vision-language models. Such robustness scalability highlights the broad applicability of TTP and its potential to serve as a lightweight yet general defense paradigm for future vision-language systems. This strong generalization capability arises from our adaptation strategy, which operates directly in the input pixel space, independent of specific text prompts or model architectures, requires neither access to model details nor any model modification, thereby enabling a truly plug-and-play defense that demonstrates strong generalization capacity.

\begin{table}[t]
\centering
\resizebox{\columnwidth}{!}{
\begin{tabular}{l|ccc|c|ccc|c}
\toprule
Method & \multicolumn{4}{c|}{Flowers} & \multicolumn{4}{c}{DTD} \\
\cmidrule(lr){2-5} \cmidrule(lr){6-9}
 & CW & DF & FGSM & Avg. & CW & DF & FGSM & Avg. \\
\midrule
CLIP~\cite{radford2021learning} & 0.8 & 0.4 & 4.8 & 2.0 & 2.3 & 7.6 & 13.4 & 7.8 \\
% Ensemble & 50.1 & 52.2 & 46.6 & 49.7 & 31.1 & 32.9 & 29.7 & 31.2 \\
% TPT~\cite{shu2022test} & 13.8 & 10.8 & 14.2 & 12.9 & 21.3 & 24.4 & 22.2 & 22.6 \\
% C-TPT~\cite{wang2023improving} & 6.6 & 5.5 & 6.2 & 6.1 & 11.9 & 15.8 & 17.5 & 15.1 \\
% MTA~\cite{zanella2024test} & 34.5 & 35.4 & 36.6 & 35.5 & 23.6 & 23.5 & 23.9 & 23.7 \\
R-TPT~\cite{sheng2025r} & 51.6 & 54.7 & 49.2 & 51.8 & 34.2 & 35.9 & 32.5 & 34.2 \\
TTP (Ours) & 
\textbf{\textcolor{cvprpink}{54.1}} & 
\textbf{\textcolor{cvprpink}{56.4}} & 
\textbf{\textcolor{cvprpink}{51.8}} & 
\textbf{\textcolor{cvprpink}{54.1}} & 
\textbf{\textcolor{cvprpink}{38.9}} & 
\textbf{\textcolor{cvprpink}{40.1}} & 
\textbf{\textcolor{cvprpink}{37.1}} & 
\textbf{\textcolor{cvprpink}{38.7}} \\
\bottomrule
\end{tabular}
}
\caption{Adversarial accuracies (\%) under CW, DeepFool (DF), and FGSM attacks on two fine-grained datasets. TTP achieves more robust performance. The best results of robustness are \textbf{\textcolor{cvprpink}{bolded}}.}
% \textbf{TTP} achieves the best robustness across all attacks.}

\label{tab:various_attacks}
\end{table}

\begin{table}[t]
\centering
\small
\resizebox{0.95\linewidth}{!}{%
\begin{tabular}{l|ccc|c}
\toprule
Method & ViT-B/32 & ViT-B/16 & ViT-L/14 & Avg. \\
\midrule
CLIP~\cite{radford2021learning} & 57.4 & 61.4 & 69.1 & 62.6 \\
TTC~\cite{xing2025clip} & 58.9 & 56.7 & 68.2 & 61.3 \\
R-TPT~\cite{sheng2025r} & 57.1 & 61.1 & 68.4 & 62.2 \\
TTP (Ours) & 57.1 & 61.2 & 68.9 & 62.4 \\
TTP (with TPT~\cite{shu2022test}) & \textbf{57.9} & \textbf{62.1} & \textbf{69.5} & \textbf{63.2} \\
\bottomrule
\end{tabular}%
}
\caption{Average clean accuracy (Acc.) of various test-time defenses across fine-grained classification datasets under various CLIP architectures. The best results of clean accuracy are \textbf{bolded}}
\label{tab:clean_acc_with_TPT}
\end{table}

% \textbf{Results on ImageNet-X.} We further evaluate our trainable TTP on ImageNet and ImageNet-OOD benchmarks. The results are summarized in Table. Overall, Trainable TTP consistently improves adversarial robustness under the ResNet-50 backbone. Notably, it achieves particularly strong performance on fine-grained classification datasets, demonstrating that our method not only performs well on ViT-based models but also exhibits strong transferability across different backbone architectures.

% \subsection{More Analysis}

\textbf{Robustness under Various Attacks.} 
To further examine the versatility and generalization of our method, we evaluate TTP and SOTA defense R-TPT under multiple adversarial attacks beyond PGD. Specifically, we employ CW~\cite{carlini2017evaluating}, DeepFool~\cite{moosavi2016deepfool}, and FGSM~\cite{goodfellow2014explaining} to assess robustness. The results on two fine-grained benchmarks are reported in Tab.~\ref{tab:various_attacks}. Across all types of attacks, TTP consistently achieves the highest adversarial accuracy, demonstrating stable and substantial gains over existing SOTA defense. These results confirm that TTP effectively mitigates a broad spectrum of adversarial attacks, establishing its versatility and attack-agnostic defense capability.

% In particular, our method has remarkable robustness in both Flower102 and DTD, which are particularly sensitive to localized texture or color perturbations. 

\textbf{Integrate with test-time adaptation.} In our proposed TTP, once an input is detected as clean, it bypasses robust adaptation and is directly forwarded for prediction. Because TTP achieves nearly 100\% detection accuracy across diverse architectures and datasets, its clean accuracy closely matches that of vanilla CLIP, effectively preserving CLIP's zero-shot generalization. Furthermore, benefiting from the detect-then-adapt strategy, samples classified as clean can be further enhanced using existing test-time adaptation methods. Taking the test-time adaptation baseline TPT~\cite{shu2022test} as an example, Tab.~\ref{tab:clean_acc_with_TPT} shows that when combined with TPT, TTP achieves the highest clean accuracy among all test-time defenses. This accuracy could even be further improved by combining TTP with stronger test-time adaptation methods. It demonstrates that TTP can be seamlessly integrated with any existing test-time adaptation method, achieving SOTA results on both clean accuracy and adversarial robustness. These results highlight the flexibility and effectiveness of our approach.
% We evaluate the performance of TTP on clean samples using the ViT-B/16 backbone on fine-grained classification datasets. For clean inputs, TTP outputs predictions directly without any modification, causing minimal impact on accuracy. The results are summarized in Tab.~\ref{tab:vitb32} and Tab.~\ref{tab:vitb16}. As shown in Tab.~\ref{tab:clean_acc_with_TPT}, when combined with test-time adaptation methods such as TPT~\cite{shu2022test}, TTP achieves the highest clean accuracy among test-time defense methods, demonstrating that it can be effectively integrated with test-time adaptation techniques without negative transfer. In contrast, R-TPT may slightly degrade clean accuracy on CLIP-ViT architectures. Overall, TTP consistently preserves clean performance while substantially improving robustness against adversarial perturbations.

\begin{figure*}[!t]
  \centering
  \begin{subfigure}[t]{0.32\textwidth}
    \centering
    \includegraphics[width=\linewidth]{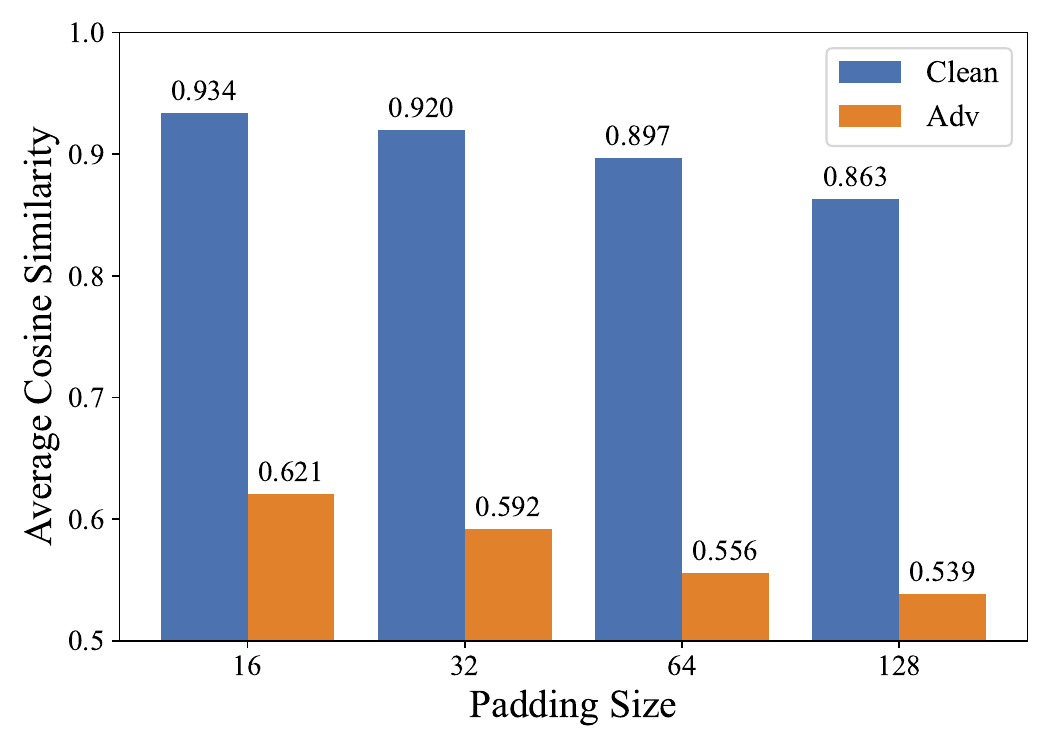}
    \caption{Average cosine similarity.}
    \label{padding_a}
  \end{subfigure}
  \hfill
  \begin{subfigure}[t]{0.32\textwidth}
    \centering
    \includegraphics[width=\linewidth]{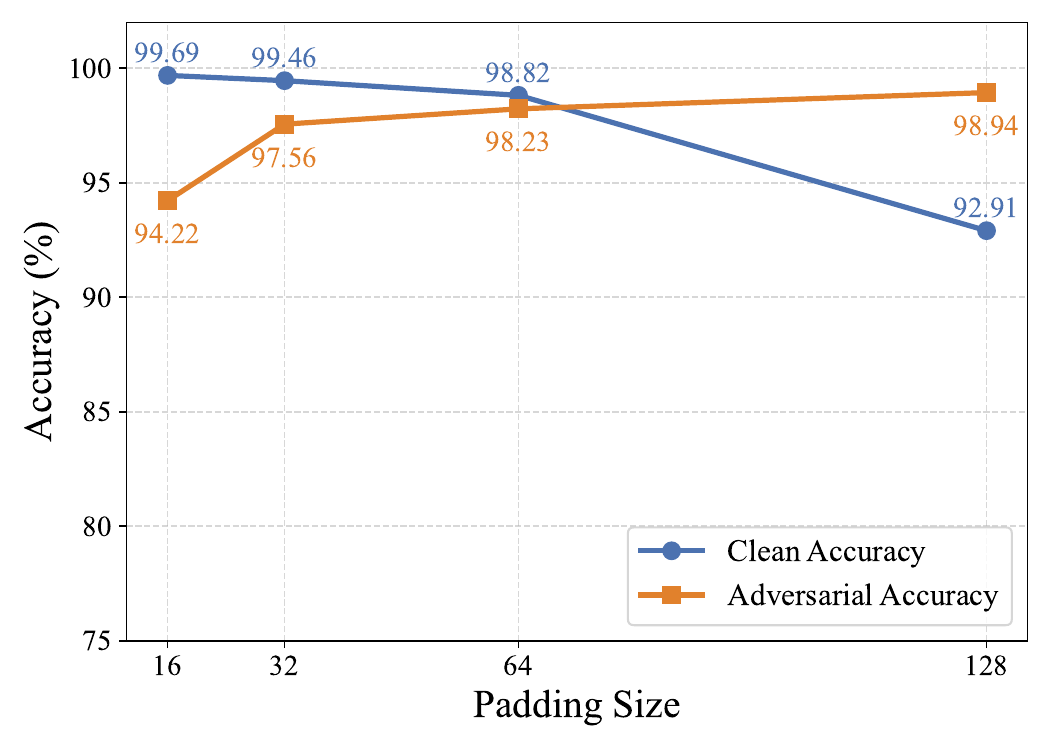}
    \caption{Detection accuracy.}
    \label{padding_b}
  \end{subfigure}
  \hfill
  \begin{subfigure}[t]{0.32\textwidth}
    \centering
    \includegraphics[width=\linewidth]{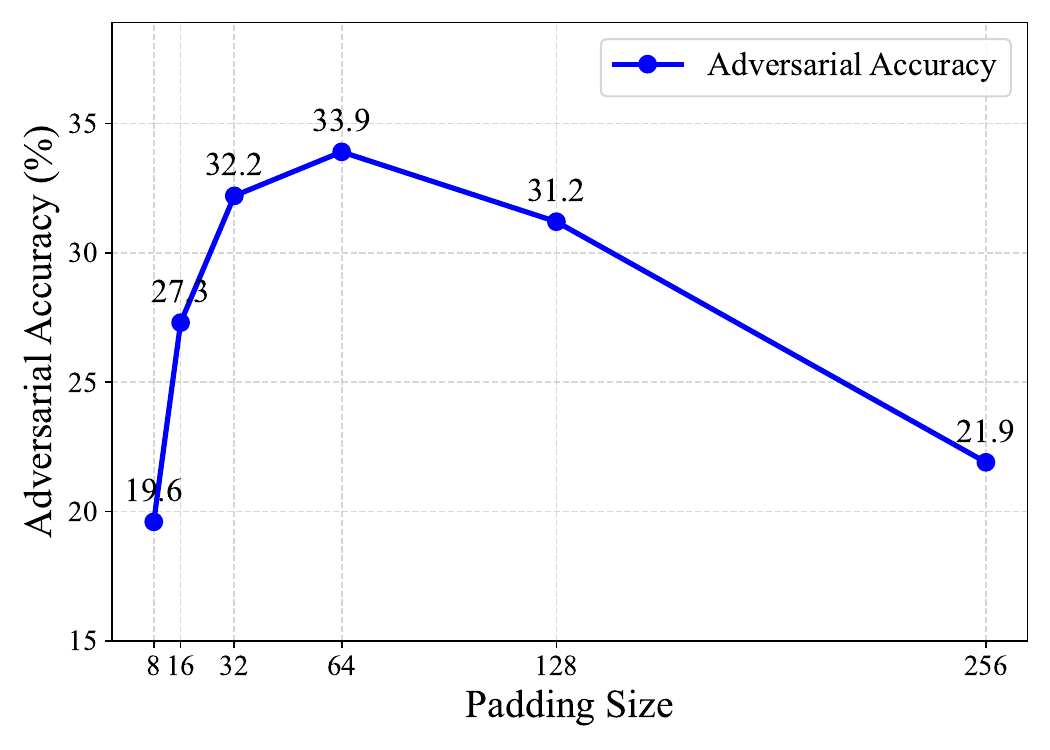}
    \caption{Adversarial Accuracy (Rob.) on DTD dataset.}
    \label{padding_c}
  \end{subfigure}
  \caption{Impact of padding size on adversarial detection and robust adaptation. ViT-B/32 is used as the CLIP backbone. The figure comprises three subplots: (a) average cosine similarities on fine-grained classification datasets of CLIP embeddings before and after padding across varying padding sizes, (b) detection accuracy for both adversarial and clean inputs, and (c) adversarial accuracy on the DTD dataset.}
  \label{fig:paddingsize}

\end{figure*}

\subsection{Ablation Study}

\textbf{Padding pattern for detection} In TTP, we detect adversarial inputs using a fixed, training-free padding scheme. We therefore examine how different padding patterns affect detection accuracy. The results in Tab.~\ref{tab:padding_type_meanacc} show that, with a predefined threshold ($\tau = 0.8$), TTP achieves consistently high accuracy on both clean and adversarial examples. Under random padding, overall detection accuracy reaches 95\%, while 0 padding (all black) and 255 padding (all white) further raise it to 98.5\% and 98.7\%, respectively. These findings indicate that simpler padding patterns yield greater accuracy gains in the detection stage; accordingly, we can adopt 0 or 255 padding for better detection.
% We evaluate the ability of TTP to detect both adversarial and clean samples across eight datasets. The results are summarized in Tab.~\ref{tab:padding_type_meanacc}. For padding size 32, under random padding, zero padding (all black), and 255 padding (all white), the feature similarity between the original and padded images exhibits substantial variation. By applying a predefined threshold ($\tau = 0.8$), TTP achieves consistently high detection accuracy for both clean and adversarial examples. In particular, under random padding, the overall detection accuracy reaches 95\%, while zero padding and 255 padding further improve the accuracy to 98.5\% and 98.7\%, respectively. Notably, adversarial examples demonstrate lower robustness under padding compared to clean samples, highlighting the effectiveness of TTP in distinguishing adversarial examples.

\textbf{Padding Size.} We conduct a systematic analysis of how padding size influences both adversarial detection accuracy and downstream robustness. For detection, we measure the cosine similarity between CLIP-encoded visual embeddings of clean samples and adversarial examples, computed before and after applying fixed zero-value spatial padding of varying sizes.
A constant detection threshold of $0.8$ is used for all settings.
As shown in Fig.~\ref{padding_a}, increasing the padding size consistently reduces the cosine similarity for adversarial examples, while inducing only minor changes for clean samples.
This widening gap between the two classes at moderate padding sizes directly improves detection accuracy by increasing separability in embedding space.
However, when padding becomes excessively large, the gap begins to narrow—likely because extreme spatial alteration disrupts the original image layout, thereby degrading both clean and adversarial feature consistency.

For robustness, we further evaluate the effect of padding size on adversarial accuracy using the DTD dataset under PGD attack.
Results in Fig.~\ref{padding_c} reveal a clear trend: performance initially rises with increasing padding size, reaching a peak at moderate values, before declining as padding grows larger.
This pattern aligns with our attention-restoration hypothesis; moderate padding appears to sufficiently mitigate adversarial effects and recover attention without compromising structural integrity, whereas overly large padding corrupts spatial context and thus undermines the model’s ability to make reliable predictions.
Taken together, these findings indicate that padding size must be carefully chosen to balance the benefits of attention restoration with the risks of structural distortion, with moderate padding emerging as the optimal setting for both high detection accuracy and strong adversarial robustness.
% We investigate the impact of padding size on adversarial detection and model robustness. For detection, we compare the cosine similarity of CLIP-encoded features for clean and adversarial examples before and after applying zero-value fixed padding under different sizes, and evaluate detection accuracy with a fixed threshold of 0.8. As shown in Fig.~\ref{padding_a}, the cosine similarity of adversarial examples decreases with increasing padding size, while that of clean samples changes only slightly, producing a growing gap and improved detection accuracy up to a moderate size. When the padding becomes too large, this gap begins to shrink as excessive padding distorts image structure and weakens feature consistency. For robustness, experiments on the DTD dataset show that adversarial accuracy first increases and then declines with larger padding, suggesting that a moderate padding size can restore disrupted attention and recover semantic information, whereas overly large padding damages visual integrity and reduces defense effectiveness.

\begin{table}[t]
\centering
\small
\begin{tabular}{l|l|l|cc}
\toprule
Sim-Aware & EntMin & Padding & ViT-B/32 & ViT-B/16 \\
\midrule
$\times$     & $\times$     & $\times$     & 0.0  & 0.0  \\
$\times$     & $\times$     & $\checkmark$ & 37.5 & 38.0 \\
$\times$     & $\checkmark$ & $\checkmark$ & 39.0 & 40.8 \\
$\checkmark$ & $\times$     & $\checkmark$ & 38.3 & 39.5 \\
$\checkmark$ & $\checkmark$ & $\checkmark$ & \textcolor{cvprpink}{\textbf{39.7}} & \textcolor{cvprpink}{\textbf{42.9}} \\
\bottomrule
\end{tabular}
\caption{\textbf{Ablation study of TTP components.} Adversarial accuracy (\%) on \textbf{fine-grained classification datasets} using ViT-B/32 and ViT-B/16 backbones.}
\label{tab:ablation}
\end{table}

\textbf{Components of TTP.} We finally conduct ablation studies on all fine-grained classification datasets using ViT-B/32 and ViT-B/16 to evaluate the contribution of each component in our proposed TTP. As shown in Tab.~\ref{tab:ablation}, introducing fixed spatial padding already substantially enhances robustness by partially restoring model attention, validating the effectiveness of our padding strategy. To further improve attention recovery, we employ entropy minimization (EntMin) to perform single-step training of the padding module during inference. Experimental results show that EntMin further boosts the model’s defensive performance. Finally, by integrating a similarity-aware ensemble, we apply additional weighting to low-entropy views for more reliable predictions. Tab.~\ref{tab:ablation} demonstrates that TTP with all components achieves the best overall performance, confirming the indispensability of each module and reinforcing the effectiveness of our approach.
% We finally conduct ablation experiments on all fine-grained classification datasets using ViT-B/32 and ViT-B/16 to assess the contribution of each component in TTP. As shown in Tab.~\ref{tab:ablation}, introducing spatial padding substantially enhances robustness by mitigating the effect of adversarial perturbations on feature alignment. 

% Incorporating entropy minimization (EntMin) during the optimization of trainable padding further stabilizes the adaptation, encouraging confident and consistent predictions under perturbations. 

% Finally, the similarity-aware ensemble, which fuses the outputs of the original and high-confidence augmented samples via similarity-based weighting, yields the best performance by integrating complementary information from multiple padded views, leading to more robust and reliable visual representations.

% \textbf{Threshold.} We further investigate the influence of the detection threshold $\tau$ on the performance of TTP. Specifically, we evaluate TTP on the DTD dataset using different CLIP backbones under varying threshold settings. As shown in the results, the choice of $\tau=0.8$ consistently achieves the best balance across different datasets and architectures. This demonstrates the strong generalization capability of our detection mechanism. In contrast, TTC\cite{xing2025clip} exhibits significant accuracy fluctuations across datasets under the same threshold setting. These results confirm that TTP provides more reliable and accurate discrimination between clean and adversarial samples compared to TTC.

\section{Conclusion}
This work introduces Test-Time Padding (TTP), a lightweight and unified defense for CLIP that achieves robust inference without retraining or architectural modification. TTP hinges on a padding-induced embedding similarity shift that reliably separates clean from adversarial inputs across various datasets and backbones, followed by targeted, single-step trainable padding and a similarity-aware ensemble to restore disrupted attention and stabilize predictions. Extensive experiments on multiple CLIP variants and fine-grained benchmarks show that TTP consistently surpasses prior test-time defenses, delivering substantial robustness gains while preserving clean accuracy. Beyond its empirical strength, TTP offers a practical blueprint for test-time defenses: detect first, then adapt, operating directly in the input space and remaining compatible with existing test-time adaptation techniques for optional clean-accuracy improvements.
% \noindent In this work, we introduce \textbf{Test-Time Padding (TTP)}, a novel adversarial defense framework that leverages adversarial example detection at test time. We begin by revisiting existing test-time defense methods and identifying their key limitations, including unstable discrimination between clean and adversarial examples and the use of uniform adaptation strategies for all inputs. Our key observation is that adversarial perturbations disrupt model attention, but this can be effectively restored through spatial padding. Building on this insight, we find that adversarial examples exhibit significantly larger representational shifts before and after padding compared to clean ones, enabling reliable detection with a single universal threshold. Guided by this detection, TTP introduces a \emph{trainable test-time padding} module and a \emph{similarity-aware ensemble} mechanism that jointly enhance robustness against adversarial attacks. Clean samples, recognized with high confidence, are directly output without adaptation, thereby preserving clean accuracy. Furthermore, TTP can be seamlessly combined with existing test-time adaptation techniques to further improve performance on benign data. Extensive experiments demonstrate that TTP consistently achieves superior adversarial robustness across diverse benchmarks and maintains or even improves clean accuracy. We believe TTP provides a new perspective on test-time defenses and contributes a step forward toward the safe deployment of vision-language models in real-world applications.

\section*{Acknowledgement}
This work was supported in part by the National Key Research and Development Program of China under Grant 2024YFC3210803, in part by the National Natural Science Foundation of China (Grant Nos. U23B2054, 62276263), and the Youth Innovation Promotion Association CAS (Grant No. Y2023143).

{
\small
\bibliographystyle{ieeenat_fullname}
\bibliography{ref2}

@String(PAMI = {IEEE Trans. Pattern Anal. Mach. Intell.})

@String(IJCV = {Int. J. Comput. Vis.})

@String(CVPR= {IEEE Conf. Comput. Vis. Pattern Recog.})

@String(ICCV= {Int. Conf. Comput. Vis.})

@String(ECCV= {Eur. Conf. Comput. Vis.})

@String(NIPS= {Adv. Neural Inform. Process. Syst.})

@String(ACMMM= {ACM Int. Conf. Multimedia})

@String(ICLR = {Int. Conf. Learn. Represent.})

@String(AAAI = {AAAI})

@String(CVPRW= {IEEE Conf. Comput. Vis. Pattern Recog. Worksh.})

@String(PAMI  = {IEEE Transactions on Pattern Analysis and Machine Intelligence})

@String(IJCV  = {International Journal of Computer Vision})

@String(CVPR  = {Proc. CVPR})

@String(ICCV  = {Proc. ICCV})

@String(ECCV  = {Proc. ECCV})

@String(NIPS  = {Proc. NeurIPS})

@String(ACMMM = {Proc. ACM MM})

@String(ICLR  = {Proc. ICLR})

@String(AAAI = {Proc. AAAI})

@String(CVPRW= {Proc. CVPR Workshops})

@String(ICCVW= {Proc. ICCV Workshops})

@String(ICML = {Proc. ICML})

@string{SP = {Proc. S\&P}}

@string{ICVGIP = {Proc. ICVGIP}}

@string{WACV = {Proc. WACV}}

@string{CoRL = {Proc. CoRL}}

@string{EMNLP = {Proc. EMNLP}}

@string{SIGIR = {Proc. ACM SIGIR}}

@inproceedings{sheng2025r,
author = {Lijun Sheng and Jian Liang and Zilei Wang and Ran He},
title = {R-TPT: Improving Adversarial Robustness of Vision-Language Models through Test-Time Prompt Tuning},
booktitle = CVPR,
pages = {29958--29967},
year = 2025
}

@inproceedings{wang2025tapt,
author = {Xin Wang and Kai Chen and Jiaming Zhang and Jingjing Chen and Xingjun Ma},
title = {Tapt: Test-time adversarial prompt tuning for robust inference in vision-language models},
booktitle = CVPR,
pages = {19910--19920},
year = 2025
}

@inproceedings{wang2024advqdet,
author = {Xin Wang and Kai Chen and Xingjun Ma and Zhineng Chen and Jingjing Chen and Yu-Gang Jiang},
title = {AdvQDet: Detecting query-based adversarial attacks with adversarial contrastive prompt tuning},
booktitle = ACMMM,
pages = {6212--6221},
year = 2024
}

@inproceedings{zhou2024few,
author = {Yiwei Zhou and Xiaobo Xia and Zhiwei Lin and Bo Han and Tongliang Liu},
title = {Few-shot adversarial prompt learning on vision-language models},
booktitle = NIPS,
volume = 37,
pages = {3122--3156},
year = 2024
}

@inproceedings{meng2025black,
author = {Fan'an Meng and Chaoran Cui and Hongjun Dai and Shuai Gong},
title = {Black-Box Test-Time Prompt Tuning for Vision-Language Models},
booktitle = AAAI,
volume = 39,
number = 6,
pages = {6099--6107},
year = 2025
}

@inproceedings{karmanov2024efficient,
author = {Adilbek Karmanov and Dayan Guan and Shijian Lu and Abdulmotaleb El Saddik and Eric Xing},
title = {Efficient test-time adaptation of vision-language models},
booktitle = CVPR,
pages = {14162--14171},
year = 2024
}

@article{ma2024imbalanced,
author = {Xingjun Ma and Linxi Jiang and Hanxun Huang and Zejia Weng and James Bailey and Yu-Gang Jiang},
title = {Imbalanced gradients: a subtle cause of overestimated adversarial robustness},
journal = {Machine Learning},
volume = 113,
number = 5,
pages = {2301--2326},
year = 2024,

}

@inproceedings{zhang2022towards,
author = {Jiaming Zhang and Qi Yi and Jitao Sang},
title = {Towards adversarial attack on vision-language pre-training models},
booktitle = ACMMM,
pages = {5005--5013},
year = 2022
}

@inproceedings{zhou2023advclip,
author = {Ziqi Zhou and Shengshan Hu and Minghui Li and Hangtao Zhang and Yechao Zhang and Hai Jin},
title = {Advclip: Downstream-agnostic adversarial examples in multimodal contrastive learning},
booktitle = ACMMM,
pages = {6311--6320},
year = 2023
}

@inproceedings{zhao2023evaluating,
author = {Yunqing Zhao and Tianyu Pang and Chao Du and Xiao Yang and Chongxuan Li and Ngai-Man Cheung and Min Lin},
title = {On evaluating adversarial robustness of large vision-language models},
booktitle = NIPS,
volume = 36,
pages = {54111--54138},
year = 2023
}

@article{madry2017towards,
author = {Aleksander Madry and Aleksandar Makelov and Ludwig Schmidt and Dimitris Tsipras and Adrian Vladu},
title = {Towards deep learning models resistant to adversarial attacks},
journal = {arXiv preprint arXiv:1706.06083},
year = 2017
}

@inproceedings{tong2025zero,
author = {Baoshun Tong and Hanjiang Lai and Yan Pan and Jian Yin},
title = {On the Zero-shot Adversarial Robustness of Vision-Language Models: A Truly Zero-shot and Training-free Approach},
booktitle = CVPR,
pages = {19921--19930},
year = 2025
}

@inproceedings{fang2024clip,
author = {Hao Fang and Jiawei Kong and Bin Chen and Tao Dai and Hao Wu and Shu-Tao Xia},
title = {Clip-guided generative networks for transferable targeted adversarial attacks},
booktitle = ECCV,
pages = {1--19},
year = 2024,

}

@inproceedings{ma2023swapprompt,
author = {Xiaosong Ma and Jie Zhang and Song Guo and Wenchao Xu},
title = {Swapprompt: Test-time prompt adaptation for vision-language models},
booktitle = NIPS,
volume = 36,
pages = {65252--65264},
year = 2023
}

@inproceedings{xing2025clip,
author = {Songlong Xing and Zhengyu Zhao and Nicu Sebe},
title = {Clip is strong enough to fight back: Test-time counterattacks towards zero-shot adversarial robustness of clip},
booktitle = CVPR,
pages = {15172--15182},
year = 2025
}

@incollection{kurakin2018adversarial,
author = {Alexey Kurakin and Ian J Goodfellow and Samy Bengio},
title = {Adversarial examples in the physical world},
booktitle = {Artificial Intelligence Safety and Security},
pages = {99--112},
year = 2018,
}

@article{fang2024one,
author = {Hao Fang and Jiawei Kong and Wenbo Yu and Bin Chen and Jiawei Li and Hao Wu and Shutao Xia and Ke Xu},
title = {One perturbation is enough: On generating universal adversarial perturbations against vision-language pre-training models},
journal = {arXiv preprint arXiv:2406.05491},
year = 2024
}

@article{he2023sa,
author = {Bangyan He and Xiaojun Jia and Siyuan Liang and Tianrui Lou and Yang Liu and Xiaochun Cao},
title = {Sa-attack: Improving adversarial transferability of vision-language pre-training models via self-augmentation},
journal = {arXiv preprint arXiv:2312.04913},
year = 2023
}

@inproceedings{lu2023set,
author = {Dong Lu and Zhiqiang Wang and Teng Wang and Weili Guan and Hongchang Gao and Feng Zheng},
title = {Set-level guidance attack: Boosting adversarial transferability of vision-language pre-training models},
booktitle = ICCV,
pages = {102--111},
year = 2023
}

@inproceedings{shu2022test,
author = {Manli Shu and Weili Nie and De-An Huang and Zhiding Yu and Tom Goldstein and Anima Anandkumar and Chaowei Xiao},
title = {Test-time prompt tuning for zero-shot generalization in vision-language models},
booktitle = NIPS,
volume = 35,
pages = {14274--14289},
year = 2022
}

@inproceedings{feng2023diverse,
author = {Chun-Mei Feng and Kai Yu and Yong Liu and Salman Khan and Wangmeng Zuo},
title = {Diverse data augmentation with diffusions for effective test-time prompt tuning},
booktitle = ICCV,
pages = {2704--2714},
year = 2023
}

@inproceedings{abdul2023align,
author = {Jameel Abdul Samadh and Mohammad Hanan Gani and Noor Hussein and Muhammad Uzair Khattak and Muhammad Muzammal Naseer and Fahad Shahbaz Khan and Salman H Khan},
title = {Align your prompts: Test-time prompting with distribution alignment for zero-shot generalization},
booktitle = NIPS,
volume = 36,
pages = {80396--80413},
year = 2023
}

@inproceedings{zanella2024test,
author = {Maxime Zanella and Ismail Ben Ayed},
title = {On the test-time zero-shot generalization of vision-language models: Do we really need prompt learning?},
booktitle = CVPR,
pages = {23783--23793},
year = 2024
}

@inproceedings{fei2004learning,
author = {Li Fei-Fei and Rob Fergus and Pietro Perona},
title = {Learning generative visual models from few training examples: An incremental bayesian approach tested on 101 object categories},
booktitle = CVPRW,
pages = {178--178},
year = 2004
}

@inproceedings{parkhi2012cats,
author = {Omkar M Parkhi and Andrea Vedaldi and Andrew Zisserman and CV Jawahar},
title = {Cats and dogs},
booktitle = CVPR,
pages = {3498--3505},
year = 2012
}

@inproceedings{krause20133d,
author = {Jonathan Krause and Michael Stark and Jia Deng and Li Fei-Fei},
title = {3d object representations for fine-grained categorization},
booktitle = ICCVW,
pages = {554--561},
year = 2013
}

@article{maji2013fine,
author = {Subhransu Maji and Esa Rahtu and Juho Kannala and Matthew Blaschko and Andrea Vedaldi},
title = {Fine-grained visual classification of aircraft},
journal = {arXiv preprint arXiv:1306.5151},
year = 2013
}

@inproceedings{cimpoi2014describing,
author = {Mircea Cimpoi and Subhransu Maji and Iasonas Kokkinos and Sammy Mohamed and Andrea Vedaldi},
title = {Describing textures in the wild},
booktitle = CVPR,
pages = {3606--3613},
year = 2014
}

@article{soomro2012dataset,
author = {Khurram Soomro and Amir Roshan Zamir and Mubarak Shah},
title = {A dataset of 101 human action classes from videos in the wild},
journal = {Center for Research in Computer Vision},
volume = 2,
number = 11,
pages = {1--7},
year = 2012
}

@article{helber2019eurosat,
author = {Patrick Helber and Benjamin Bischke and Andreas Dengel and Damian Borth},
title = {Eurosat: A novel dataset and deep learning benchmark for land use and land cover classification},
journal = {IEEE Journal of Selected Topics in Applied Earth Observations and Remote Sensing},
volume = 12,
number = 7,
pages = {2217--2226},
year = 2019,

}

@inproceedings{nilsback2008automated,
author = {Maria-Elena Nilsback and Andrew Zisserman},
title = {Automated flower classification over a large number of classes},
booktitle = ICVGIP,
pages = {722--729},
year = 2008,

}

@inproceedings{moosavi2016deepfool,
author = {Seyed-Mohsen Moosavi-Dezfooli and Alhussein Fawzi and Pascal Frossard},
title = {Deepfool: a simple and accurate method to fool deep neural networks},
booktitle = CVPR,
pages = {2574--2582},
year = 2016
}

@inproceedings{jia2021scaling,
author = {Chao Jia and Yinfei Yang and Ye Xia and Yi-Ting Chen and Zarana Parekh and Hieu Pham and Quoc Le and Yun-Hsuan Sung and Zhen Li and Tom Duerig},
title = {Scaling up visual and vision-language representation learning with noisy text supervision},
booktitle = ICML,
pages = {4904--4916},
year = 2021,
}

@article{huang2023visual,
author = {Zhi Huang and Federico Bianchi and Mert Yuksekgonul and Thomas J Montine and James Zou},
title = {A visual--language foundation model for pathology image analysis using medical twitter},
journal = {Nature Medicine},
volume = 29,
number = 9,
pages = {2307--2316},
year = 2023,

}

@inproceedings{wang2022medclip,
author = {Zifeng Wang and Zhenbang Wu and Dinesh Agarwal and Jimeng Sun},
title = {Medclip: Contrastive learning from unpaired medical images and text},
booktitle = EMNLP,
volume = 2022,
pages = {3876},
year = 2022
}

@article{ahn2022can,
author = {Michael Ahn and Anthony Brohan and Noah Brown and Yevgen Chebotar and Omar Cortes and Byron David and Chelsea Finn and Chuyuan Fu and Keerthana Gopalakrishnan and Karol Hausman and others},
title = {Do as i can, not as i say: Grounding language in robotic affordances},
journal = {arXiv preprint arXiv:2204.01691},
year = 2022
}

@inproceedings{shridhar2022cliport,
author = {Mohit Shridhar and Lucas Manuelli and Dieter Fox},
title = {Cliport: What and where pathways for robotic manipulation},
booktitle = CoRL,
pages = {894--906},
year = 2022,

}

@article{goodfellow2014explaining,
author = {Ian J Goodfellow and Jonathon Shlens and Christian Szegedy},
title = {Explaining and harnessing adversarial examples},
journal = {arXiv preprint arXiv:1412.6572},
year = 2014
}

@inproceedings{sui2025just,
author = {Elaine Sui and Xiaohan Wang and Serena Yeung-Levy},
title = {Just shift it: Test-time prototype shifting for zero-shot generalization with vision-language models},
booktitle = WACV,
pages = {825--835},
year = 2025,

}

@article{liang2025comprehensive,
author = {Jian Liang and Ran He and Tieniu Tan},
title = {A comprehensive survey on test-time adaptation under distribution shifts},
journal = IJCV,
volume = 133,
number = 1,
pages = {31--64},
year = 2025,

}

@inproceedings{wu2020adversarial,
author = {Dongxian Wu and Shu-Tao Xia and Yisen Wang},
title = {Adversarial weight perturbation helps robust generalization},
booktitle = NIPS,
volume = 33,
pages = {2958--2969},
year = 2020
}

@inproceedings{zhang2019theoretically,
author = {Hongyang Zhang and Yaodong Yu and Jiantao Jiao and Eric Xing and Laurent El Ghaoui and Michael Jordan},
title = {Theoretically principled trade-off between robustness and accuracy},
booktitle = ICML,
pages = {7472--7482},
year = 2019,

}

@inproceedings{zhang2024universal,
author = {Peng-Fei Zhang and Zi Huang and Guangdong Bai},
title = {Universal adversarial perturbations for vision-language pre-trained models},
booktitle = SIGIR,
pages = {862--871},
year = 2024
}

@inproceedings{wang2024transferable,
author = {Haodi Wang and Kai Dong and Zhilei Zhu and Haotong Qin and Aishan Liu and Xiaolin Fang and Jiakai Wang and Xianglong Liu},
title = {Transferable multimodal attack on vision-language pre-training models},
booktitle = SP,
pages = {1722--1740},
year = 2024,

}

@inproceedings{carlini2017evaluating,
author = {Nicholas Carlini and David Wagner},
title = {Towards evaluating the robustness of neural networks},
booktitle = SP,
year = 2017
}

@article{chen2023vlp,
author = {Fei-Long Chen and Du-Zhen Zhang and Ming-Lun Han and Xiu-Yi Chen and Jing Shi and Shuang Xu and Bo Xu},
title = {Vlp: A survey on vision-language pre-training},
journal = {Machine Intelligence Research},
volume = 20,
number = 1,
pages = {38--56},
year = 2023
}

@article{hendrycksaugmix,
  title={Augmix: A simple data processing method to improve robustness and uncertainty},
  author={Hendrycks, Dan and Mu, Norman and Cubuk, Ekin D and Zoph, Barret and Gilmer, Justin and Lakshminarayanan, Balaji},
  journal={arXiv preprint arXiv:1912.02781},
  year={2019}
}

@article{kong2024patch,
author = {Dehong Kong and Siyuan Liang and Xiaopeng Zhu and Yuansheng Zhong and Wenqi Ren},
title = {Patch is enough: naturalistic adversarial patch against vision-language pre-training models},
journal = {Visual Intelligence},
volume = 2,
number = 1,
pages = {1--10},
year = 2024
}

@inproceedings{li2024one,
author = {Lin Li and Haoyan Guan and Jianing Qiu and Michael Spratling},
title = {One prompt word is enough to boost adversarial robustness for pre-trained vision-language models},
booktitle = CVPR,
year = 2024
}

@inproceedings{mao2023understanding,
author = {Chengzhi Mao and Scott Geng and Junfeng Yang and Xin Wang and Carl Vondrick},
title = {Understanding zero-shot adversarial robustness for large-scale models},
booktitle = ICLR,
year = 2023
}

@inproceedings{radford2021learning,
author = {Alec Radford and Jong Wook Kim and Chris Hallacy and Aditya Ramesh and Gabriel Goh and Sandhini Agarwal and Girish Sastry and Amanda Askell and Pamela Mishkin and Jack Clark and others},
title = {Learning transferable visual models from natural language supervision},
booktitle = ICML,
year = 2021
}

@inproceedings{szegedy2014intriguing,
author = {Christian Szegedy and Wojciech Zaremba and Ilya Sutskever and Joan Bruna and Dumitru Erhan and Ian Goodfellow and Rob Fergus},
title = {Intriguing properties of neural networks},
booktitle = ICLR,
year = 2014
}

@inproceedings{wang2021tent,
author = {Dequan Wang and Evan Shelhamer and Shaoteng Liu and Bruno Olshausen and Trevor Darrell},
title = {Tent: Fully test-time adaptation by entropy minimization},
booktitle = ICLR,
year = 2021
}

@article{yu2023benchmarking,
author = {Yongcan Yu and Lijun Sheng and Ran He and Jian Liang},
title = {Benchmarking test-time adaptation against distribution shifts in image classification},
journal = {arXiv preprint arXiv:2307.03133},
year = 2023
}

@article{zhang2024vision,
author = {Jingyi Zhang and Jiaxing Huang and Sheng Jin and Shijian Lu},
title = {Vision-language models for vision tasks: A survey},
journal = PAMI,
year = 2024
}
}
\end{document}